\documentclass{article}

\usepackage[final]{nips_2016}

\usepackage[utf8]{inputenc} 
\usepackage[T1]{fontenc}    
\usepackage{hyperref}       
\usepackage{url}            
\usepackage{booktabs}       
\usepackage{amsfonts}       
\usepackage{nicefrac}       
\usepackage{microtype}      

\usepackage{amsmath}
\usepackage{amsthm}
\usepackage{amssymb}
\usepackage{MnSymbol}
\usepackage{mathtools}
\usepackage{algorithm}
\usepackage{algpseudocode}
\usepackage{caption}
\makeatletter
\@addtoreset{algorithm}{section}
\makeatother

\newcommand{\R}[1]{\mathbb{R}^{#1}}
\newcommand{\lin}[2]{\mathcal{L}(#1; #2)}
\newcommand{\bigcrop}{\mathcal{K}}
\newcommand{\smallcrop}{\kappa}
\newcommand{\bigpool}{\Psi}
\newcommand{\smallpool}{\psi}
\newcommand{\lefthook}{\righthalfcup}
\newcommand{\righthook}{\lefthalfcup}
\newcommand{\bignl}{S}
\newcommand{\smallnl}{\sigma}
\newcommand{\ip}[2]{\langle #1, \,#2 \rangle}

\newcommand{\Em}{\operatorname{Em}}
\newcommand{\dd}[2]{\frac{\mathrm{d} #1}{\mathrm{d} #2}}
\newcommand{\D}{\mathrm{D}}
\newcommand{\eval}[1]{\bigg|_{#1}}
\newcommand{\norm}[1]{\left\|#1\right\|}
\newcommand{\head}{\alpha}
\newcommand{\tail}{\omega}

\newtheorem{theorem}{Theorem}
\newtheorem{lemma}[theorem]{Lemma}

\title{A Geometric Framework for \\ Convolutional Neural Networks}

\author{
  Anthony L. Caterini and Dong Eui Chang \\
  Department of Applied Mathematics \\
  University of Waterloo \\
  Waterloo, ON, Canada, N2L 3G1 \\
  \texttt{\{alcateri, dechang\}@uwaterloo.ca}
}

\begin{document}

\maketitle

\begin{abstract}
In this paper, a geometric framework for neural networks is proposed. This framework uses the inner product space structure underlying the parameter set to perform gradient descent not in a component-based form, but in a coordinate-free manner. Convolutional neural networks  are described in this framework in a compact form, with the gradients of standard --- and higher-order --- loss functions calculated for each layer of the network. This approach can be applied to other network structures and provides a basis on which to create new networks. 
\end{abstract}

\section{Introduction}
Machine Learning algorithms have long worked with multi-dimensional vector data and parameters, but have not exploited the underlying inner product space structure. A recent paper on deep learning in \emph{Nature} called for ``new paradigms'' involving ``operations on large vectors'' \cite{lecun2015deep} to propel the field forward. This approach is taken to describe the convolutional neural network (CNN) in this paper. In particular, the layers are described as vector-valued maps, and gradients of these maps with respect to the parameters at each layer are taken in a coordinate-free manner. This approach promotes a greater understanding of the network than a coordinate-based approach, and allows for loss function gradients to be calculated compactly using coordinate-free backpropagation of error. This paper also considers a higher-order loss function, as in \cite{rifai2011manifold} and \cite{simard1992tangent}. Algorithms to compute one iteration of gradient descent are provided for both types of loss functions to clarify the application of the developed theory. The precise notation developed throughout this paper provides a mathematical standard upon which deep learning can be researched, overcoming the inconsistent notation currently employed across the field. The framework developed in this paper is flexible, and can be extended to cover other types of network structures, and even inspire further developments in deep learning. 

\section{Mathematical Preliminaries} 
Some prerequisite notation and concepts are introduced here before CNNs can be fully described.

\subsection{Multilinear Algebra and  Derivatives}
Every individual vector space is assumed to be an inner product space, with the inner product represented by $\ip{\,}{}$. The inner product naturally extends to the direct product  $E_1 \times \cdots \times E_r$ of inner product spaces $E_1, \ldots, E_r$ and their tensor product  $E_1 \otimes \cdots \otimes E_r$ as follows  \cite{werner1978multilinear}:
\begin{align*}
\ip{(e_1, \cdots, e_r)}{(\bar e_1, \cdots, \bar e_r)} = \sum_{i=1}^r \ip{e_i}{\bar e_i}, \quad \ip{e_1 \otimes \cdots \otimes e_r}{\bar e_1\otimes \cdots \otimes \bar e_r} = \prod_{i=1}^r \ip{e_i}{\bar e_i},
\end{align*}
where $e_i, \bar e_i \in E_i$, $i =1, \ldots, r$. The symbol $\otimes$ is exclusively used to denote the tensor product operator in this paper. An inner product space $E$ is canonically identified here with its dual space $E^*$ using the inner product on $E$, so dual spaces will rarely be used in this paper. The set of $r$-linear maps from  $E_1  \times   \cdots \times E_r$ to a vector space $F$ is denoted by $\mathcal L(E_1,  \ldots, E_r; F)$. For a linear map $L \in \lin{E}{F}$, its adjoint map, denoted by $L^*$, is a linear map in $\lin{F}{E}$  defined by the relationship  $\ip{L^* f}{e} = \ip{f}{L e}$ for all $e\in E$ and $f\in F$. For each vector $e_1 \in E_1$ and any bilinear map $B \in \lin{E_1,E_2}{F}$, define a linear map $(e_1 \lefthook B): E_2 \rightarrow F$ by 
\[
(e_1 \lefthook B ) (e) = B(e_1,e)
\]
for all $e\in E_2$. Likewise, for each vector $e_2 \in E_2$ and any bilinear map $B \in \lin{E_1,E_2}{F}$,  define a linear map $( B \righthook e_2): E_1 \rightarrow F$ by 
\[
( B \righthook e_2) (e) = B(e,e_2)
\]
for all $e\in E_1$.

Now, notation for derivatives in accordance with \cite{marsden1988manifolds} is presented. Consider a map $f: E_1 \rightarrow  E_2$. The (first) derivative $\D f(x)$ of $f$ at a point $x \in E_1$ is a linear map from $E_1$ to $E_2$, i.e. $\D f(x) \in \lin{E_1}{E_2}$, and it can be defined as 
\[
\D f(x) \cdot v = \left . \dd{}{t} f(x+tv)\right |_{t=0}
\]
for any $v \in E_1$. The derivative $\D f$ can be viewed as a map from $E_1$ to $\lin{E_1}{E_2}$, defined by $x \mapsto \D f(x)$. Let  $\D^*f(x)$ denote the adjoint of $\D f(x)$ so that $\ip{w}{\D f(x) \cdot v} = \ip{\D^*f(x) \cdot w}{v}$ for all $v \in E_1$ and $w \in E_2$. 

Now consider a map $f:  E_1 \times F_1 \rightarrow E_2$ written as $f(x;\theta)$ for $x \in E_1$ and $\theta \in F_1$, where the semi-colon is inserted between $x$ and $\theta$ to distinguish the state variable $x$ from the parameters $\theta$. Let $\D f(x;\theta)$ denote the derivative of $f$ with respect to $x$ evaluated at $(x;\theta)$, and let  $\nabla f (x;\theta)$ denote the derivative of $f$ with respect to $\theta$ evaluated at $(x;\theta)$.  It is easy to verify that $\D f(x;\theta) \in \mathcal{L}(E_1; E_2)$ and $\nabla f(x;\theta) \in \mathcal{L}(F_1; E_2)$ and that 
\[
\D f(x;\theta) \cdot e =  \left . \dd{}{t} f(x + t e;\theta) \right |_{t=0}, \qquad \nabla f(x;\theta)\cdot u =  \left . \dd{}{t} f(x; \theta + t u) \right |_{t=0}
\]
 for all $e \in E_1$ and  $u \in F_1$.  The adjoints  of $\D f(x;\theta)$ and $\nabla f(x;\theta)$ are denoted by $\D^*f(x;\theta)$ and $\nabla^*f(x;\theta)$, respectively. Sometimes,  $\nabla_\theta f$ is written instead of $\nabla f$, to emphasize differentiation of $f$ with respect to the parameter variable $\theta$. 

The second derivative $\D^2f(x;\theta)$ of $f$ with respect to  $x$ evaluated at $(x;\theta)$ is a bilinear map in $\mathcal L(E_1, E_1; E_2)$ defined as follows: for any $e, \bar e \in E_1$, 
\[
\D^2f(x;\theta)\cdot (e,\bar e) =  \left. \dd{}{t}\D f(x+te;\theta)\cdot \bar e \, \right  |_{t=0} = \left. \frac{\partial^2}{\partial t \partial s} f(x + te + s \bar{e}; \theta) \right|_{t = s = 0}.
\]
It is assumed that every function that appears in this paper is (piecewise) twice continuously differentiable. 
The second derivative $\D^2f (x;\theta)$ is symmetric, i.e. $\D^2f(x;\theta)\cdot (e,\bar e) = \D^2f(x;\theta)\cdot (\bar e, e)$ for all $e, \bar e \in E_1$. The second derivative $\D\nabla f(x;\theta)$ of $f$ with respect to $x$ and $\theta$ at the point $(x;\theta)$ is a bilinear map in $\mathcal L (E_1, F_1; E_2)$ defined as follows: for any $e \in E_1$ and $u \in F_1$, 
 \[
 \D\nabla f(x;\theta) \cdot  (e,u) = \left .  \dd{}{t} \nabla f(x+t e; \theta)\cdot u \, \right |_{t=0}.
 \]
  On the other hand, the second derivative $\nabla \D f(x;\theta)$ of $f$ with respect to $\theta$ and $x$ at the point $(x;\theta)$ denotes the bilinear map in $\mathcal L (F_1, E_1; E_2)$ defined as follows: for any $u \in F_1$ and $e \in E_1$,  
 \[
\nabla \D f(x;\theta) \cdot  (u,e) = \left.  \dd{}{t} \D f(x; \theta+tu)\cdot e \, \right |_{t=0}.
\]
Note that for all $e\in E_1$ and $u \in F_1$, it is easy to verify that
\begin{align}\label{mixed:partials}
\D \nabla f(x;\theta)\cdot (e,u) = \nabla \D f(x;\theta)\cdot (u,e).
\end{align}

\subsection{Backpropagation in a Nutshell}
Now, backpropagation will be presented in a coordinate-free form. Given two maps $f_1(x;\theta_1) \in E_2$ for $x \in E_1, \theta_1 \in F_1$ and $f_2(z;\theta_2) \in E_3$ for $z \in E_2, \theta_2 \in F_2$, the composition $f:= f_2\circ f_1$ is the map defined as follows: 
\begin{align} \label{eqn:f_comp}
f(x; \theta_1, \theta_2) = f_2(f_1(x;\theta_1); \theta_2),
\end{align} 
for $x\in E_1, \theta_1 \in F_1, \theta_2 \in F_2$. In this framework, functions are composed with respect to the state variables. By the chain rule, 
\begin{align}\label{chain:rule}
\D f = (\D f_2 \circ f_1) \cdot \D f_1, \quad \nabla_{\theta_1}f = (\D f_2 \circ f_1 ) \cdot \nabla_{\theta_1}f_1, \quad \nabla_{\theta_2}f = \nabla_{\theta_2}f_2 \circ f_1,
\end{align}
which are evaluated at a point $(x; \theta_1, \theta_2)$ as follows:
\begin{align*}
\D f(x) = \D f_2 (f_1(x))\cdot \D f_1(x), \quad \nabla_{\theta_1}f = \D f_2 (f_1 (x)) \cdot \nabla_{\theta_1}f_1(x), \quad \nabla_{\theta_2}f (x)= \nabla_{\theta_2}f_2 (f_1(x)),
\end{align*}
where the dependency on the parameters $\theta_1$ and $ \theta_2$ is suppressed for brevity, which shall be understood throughout the paper. 
In particular, taking the adjoint of  $\nabla_{\theta_1}f = (\D f_2 \circ f_1 ) \cdot \nabla_{\theta_1}f_1$ produces
\begin{align}\label{backprop:nutshell}
\nabla_{\theta_1}^* f  = ((\D f_2 \circ f_1 ) \cdot \nabla_{\theta_1}f_1)^* = \nabla_{\theta_1}^* f_1 \cdot (\D^*f_2 \circ f_1)
\end{align}
which is \emph{backpropagation} in a nutshell. This can be seen by the following: consider a loss function $J$ defined by
\begin{align*}
J(x; \theta_1, \theta_2) = \frac{1}{2} \norm{f(x; \theta_1, \theta_2) -  y}^2 = \frac{1}{2} \ip{f(x; \theta_1, \theta_2) -  y}{f(x; \theta_1, \theta_2) -  y}
\end{align*}
for some vector $y \in E_3$ that may depend on $x$, along with $f$ as in (\ref{eqn:f_comp}). Then, for any $u \in F_1$, with $\theta \coloneqq \{\theta_1, \theta_2\}$ representing the parameters, 
\begin{align} \label{eqn:nabla_J_u}
\nabla_{\theta_1} J(x;\theta) \cdot u = \ip{f(x; \theta) - y}{\nabla_{\theta_1}f(x; \theta) \cdot u} = \ip{\nabla^*_{\theta_1}f(x; \theta) \cdot (f(x; \theta) - y)}{u}.
\end{align}
Since this holds for any $u \in F_1$,  the canonical  identification of an inner product space with its dual is used to obtain
\begin{align} \label{eqn:nabla_J}
\nabla_{\theta_1}J(x;\theta) &= \nabla^*_{\theta_1}f(x; \theta)\cdot ( f(x; \theta) - y) \\
&= \nabla_{\theta_1}^* f_1(x;\theta_1) \cdot \D^*f_2 (f_1(x;\theta_1);\theta_2) \cdot ( f(x; \theta) -  y), \nonumber
\end{align}
where   (\ref{backprop:nutshell}) is used for the second equality. This shows that the error $(f(x; \theta) - y)$ propagates backward from layer 2 to layer 1 through multiplication by $\D^*f_2$. The adjoint operator reverses the direction of composition, i.e. $(L_1 L_2)^* = L_2^* L_1^*$, which is the key to backpropagating the error. 

The second derivative $\D^2f$ of $f = f_2 \circ f_1$ is given by
\[
\D^2 f(x) \cdot (e,\bar e) = \D^2 f_2(f_1(x)) \cdot (\D f_1(x) \cdot e, \D f_1(x) \cdot \bar e) + \D f_2(f_1(x)) \cdot \D^2f_1(x)\cdot (e,\bar e)
\]
for all $e,\bar e\in E_1$. The second derivative $\nabla_{\theta_1} \D f$ is given by
\begin{align*}
\nabla_{\theta_1}\D f (x)  \cdot (u,e) =\!  \D^2 f_2 (f_1(x)) \cdot (\nabla_{\theta_1} f_1(x) \cdot u, \D f_1(x) \cdot e) \! + \! \D f_2(f_1(x)) \cdot \nabla_{\theta_1} \D f_1(x) \cdot (u,e)
\end{align*}
for all $e\in E_1$ and  $u\in F_1$, which is equivalent to the following: for  any fixed  $e\in E_1$
\begin{equation}\label{backprop2:NabD}
\nabla_{\theta_1}\D f (x) \righthook e = (\D^2 f_2 (f_1(x)) \righthook (\D f_1(x) \cdot e) ) \cdot \nabla_{\theta_1} f_1(x) + \D f_2(f_1(x)) \cdot (\nabla_{\theta_1} \D f_1(x) \righthook e),
\end{equation}
which is a linear map from $F_1$ to $E_3$, or by (\ref{mixed:partials})
\begin{equation}\label{backprop2:DNab}
e \lefthook \D \nabla_{\theta_1} f (x)  = \left(( \D f_1(x) \cdot e ) \lefthook \D^2 f_2 \left(f_1(x)\right) \right) \cdot \nabla_{\theta_1} f_1(x) + \D f_2(f_1(x)) \cdot (e \lefthook \D \nabla_{\theta_1} f_1(x) ).
\end{equation}
The adjoint of (\ref{backprop2:NabD}) or  (\ref{backprop2:DNab}) yields higher-order backpropagation of error, say for a loss function $R = \frac{1}{2} \| \D f(x)\cdot e - y\|^2$  with some $e\in E_1$ and $y \in E_3$ that may depend on $x$, but not on the parameters. Higher-order backpropagation will be studied in more detail  in the next section.

Backpropagation  can be  expressed  recursively for the composition of more than two functions. Consider $L$ functions $f_t(x_t; \theta_t) \in E_{t+1}$ for $x_t \in E_t, \theta_t \in F_t$, $t=1, \ldots, L $. Define the composition $F \coloneqq f_L\circ \cdots \circ f_1$. Let $\omega_t = f_L \circ \cdots \circ f_t$ and $\head_t = f_t \circ \cdots \circ f_1$ for $t=1,\ldots, L$ so that 
\begin{equation}\label{eqn:recursive_rel}
F = \omega_{t+1} \circ \head_t, \quad \omega_t = \omega_{t+1} \circ f_{t}, \quad \head_{t+1} = f_{t+1} \circ \head_t
\end{equation}
 for all $t = 1, \ldots, L-1$.  The first and second derivatives of  (\ref{eqn:recursive_rel}) and their adjoints can be easily obtained.
 
\section{Convolutional Neural Networks}
This section will describe how the above framework can be applied to convolutional neural networks; refer to \cite{huang2006large} or \cite{krizhevsky2012imagenet}, for example, for more on the theory of CNNs. First, the actions of one layer of a generic CNN will be described, and then this will be extended to multiple layers. A coordinate-free gradient descent algorithm will also be described. Note that in this section, all bases of inner product space will be assumed to be orthonormal.

\subsection{Single Layer Formulation}
The actions of one layer of the network will be denoted $f(X; W,B)$, where $X \in \R{n_1 \times \ell_1} \otimes \R{m_1}$ is the state variable, and  $W \in \R{p \times q} \otimes \R{m_2}$ and $B \in \R{\bar{n}_1 \times \bar{\ell}_1} \otimes \R{m_2}$ are the parameters. Throughout this section, let $\{e_i\}_i$ (resp. $\{\tilde{e}_a\}_a$) be a basis for $\R{m_1}$ (resp. $\R{m_2}$), and let $\{E_{jk}\}_{jk}$ (resp. $\{\bar{E}_{jk}\}_{jk}, \{\tilde{E}_{jk}\}_{jk}, \{\hat{E}_{jk}\}_{jk}$) be a basis for $\R{n_1 \times \ell_1}$ (resp. $\R{\bar{n}_1 \times \bar{\ell}_1}, \R{p \times q}, \R{n_2 \times \ell_2}$). Then $X$, $W$ and $B$ can be written as follows:
\begin{align*}
X = \sum_{i=1}^{m_1} X_i \otimes e_i, \quad W = \sum_{a=1}^{m_2} W_a \otimes \tilde{e}_a, \quad B = \sum_{a=1}^{m_2} B_a \otimes \tilde{e}_a.
\end{align*}
Each $X_i \in \R{n_1 \times \ell_1}$ is called a \emph{feature map}, which corresponds to an abstract representation of the input for a generic layer. Each $W_a \in \R{p \times q}$ is a \emph{filter} used in convolution, and each $B_a \in \R{\bar{n}_1 \times \bar{\ell}_1}$ is a \emph{bias} term. The actions of the layer are then a new set of feature maps, $f(X; W, B) \in \R{n_2 \times \ell_2} \otimes \R{m_2}$, with explicit form given by:
\begin{align} \label{eqn:single_layer}
f(X; W, B) = \bigpool\left(\bignl(C(W,X) + B)\right),
\end{align}
where  $\bigpool$ is a pooling operator, $\bignl$ is an elementwise nonlinear function, and $C$ is the convolution operator, all of which will be defined in this section. 

\subsubsection{Cropping, Embedding and Mixing Operators}
The \emph{cropping} and \emph{mixing} operators will be used to define the convolution operator $C$ that appears in (\ref{eqn:single_layer}). The cropping operator, $\bigcrop_{jk} \in \lin{\R{n_1 \times \ell_1} \otimes \R{m_1}}{\R{p \times q} \otimes \R{m_1}}$, is defined as:
\begin{align} \label{eqn:bigcrop}
\bigcrop_{jk}\left(\sum_{i=1}^{m_1} X_i \otimes e_i\right) &\coloneqq \sum_{i=1}^{m_1} \smallcrop_{jk}(X_i) \otimes e_i,
\end{align}
where $\smallcrop_{jk} \in \lin{\R{n_1 \times \ell_1}}{\R{p \times q}}$ is defined as:
\begin{align} \label{eqn:smallcrop}
\smallcrop_{jk}(X_i) \coloneqq \sum_{r=1}^p \sum_{s=1}^q \ip{X_i}{ E_{j+r-1, k+s-1}} \tilde{E}_{rs}.
\end{align}
Define the \emph{embedding} operator $\Em_{c,d} \in \lin{\R{p \times q}}{\R{n_1 \times \ell_1}}$ by
\begin{align} \label{eqn:embed}
\Em_{c,d}(Y) = \sum_{r = 1}^p \sum_{s = 1}^q Y_{rs} E_{c + r-1,d + s -1}
\end{align}
 for $Y = \sum_{r=1}^p \sum_{s=1}^q Y_{rs} \tilde{E}_{rs} \in \R{p \times q}$, 
which corresponds to embedding $Y$ into the zero matrix when  $\{E_{jk}\}_{jk}$ is the standard basis. The adjoints of $\bigcrop_{jk}$ and  $\smallcrop_{jk}$ are calculated as follows:
\begin{theorem}
For any $Z = \sum_{i=1}^{m_1} Z_i \otimes {e}_i \in \R{p \times q} \otimes \R{m_1}$, 
\begin{align*}
\bigcrop^*_{jk} (Z) = \sum_{i=1}^{m_1}\smallcrop^*_{jk}(Z_i) \otimes e_i,
\end{align*}
where, for any $i \in \{1, \ldots, m_1\}$, 
\begin{equation}\label{small_kappa_star}
\smallcrop_{jk}^* (Z_i) = \Em_{j,k}(Z_i).
\end{equation}
\begin{proof} Let $Q \in \R{n_1 \times \ell_1}$.  Then, for any $i \in \{1, \ldots, m_1\}$, 
\begin{align*}
\ip{Z_i}{\smallcrop_{jk} (Q)} &= \left\langle Z_i, \sum_{r=1}^p \sum_{s=1}^q \ip{Q}{E_{j+r-1, k+s-1}} \tilde{E}_{rs} \right\rangle \\
&= \left\langle \sum_{r=1}^p \sum_{s=1}^q \ip{Z_i}{\tilde{E}_{rs}} E_{j+r-1, k+s-1}, Q\right\rangle \\
&= \ip{\Em_{j,k}(Z_i)}{Q},
\end{align*}
which proves \eqref{small_kappa_star}.
Furthermore, let $X = \sum_{i=1}^{m_1} X_i \otimes e_i \in \R{n_1 \times \ell_1} \otimes \R{m_1}$. Then, 
\begin{align*}
\ip{Z}{\bigcrop_{jk} (X)} &= \left\langle \sum_{i=1}^{m_1} Z_i \otimes e_i, \sum_{i=1}^{m_1} \smallcrop_{jk} (X_i) \otimes e_i\right\rangle \\
&= \sum_{i=1}^{m_1} \ip{Z_i}{\smallcrop_{jk}(X_i)} \\
&= \sum_{i=1}^{m_1} \ip{\smallcrop^*_{jk} (Z_i)}{X_i} \\
&= \left\langle\sum_{i=1}^{m_1} \smallcrop^*_{jk} (Z_i) \otimes e_i, X\right\rangle,
\end{align*}
which completes the proof
\end{proof}
\end{theorem}
For $v \in \R{m_1}$, the \emph{mixing} operator $\Phi_v \in \lin{\R{p \times q} \otimes \R{m_1}}{\R{p \times q}}$ defines how the cropped feature maps are combined into the next layer of feature maps, which is useful in a framework such as \cite{lecun1998gradient}. It can be explicitly represented as:
\begin{align} \label{eqn:Phi}
\Phi_v\left(\sum_{i=1}^{m_1} U_i \otimes e_i\right) = \sum_{i=1}^{m_1} v_i U_i,
\end{align}
where $v = \sum_{i=1}^{m_1} v_i e_i$. The adjoint operator $\Phi^*_v$ has a compact form, as the following lemma describes.
\begin{lemma}
For any $Y \in \R{p \times q}$ and $v \in \R{m_1}$,
\begin{align*}
\Phi_v^* \cdot Y = Y \otimes v. 
\end{align*}
\begin{proof} Let $X = \sum_{i=1}^{m_1} X_i \otimes e_i \in \R{p \times q} \otimes \R{m_1}$. Then, 
\begin{align*}
\ip{Y}{\Phi_v (X)} &= \left\langle Y, \sum_{i=1}^{m_1} v_i X_i\right\rangle \\
&= \sum_{i=1}^{m_1} \ip{v_i Y}{X_i} \\
&= \left\langle \sum_{i=1}^{m_1} \left(v_i Y\right) \otimes e_i,X\right\rangle\\
&= \left\langle Y \otimes \left(\sum_{i=1}^{m_1} v_i e_i\right),X\right\rangle \\
&= \ip{Y \otimes v}{X}.
\end{align*}
This implies that $\Phi^*_v \cdot Y = Y \otimes v$ since the above equations are true for any $X$. 
\end{proof}
\end{lemma}

\subsubsection{Convolution Operator}
The $C$ operator in (\ref{eqn:single_layer}) is known as the \emph{convolution} operator. The convolution $C \in \lin{\R{p \times q} \otimes \R{m_2}, \R{n_1 \times \ell_1} \otimes \R{m_1}}{\R{\bar{n}_1 \times \bar{\ell}_1} \otimes \R{m_2}}$ is defined as:
\begin{align*}
C(W,X) = \sum_{a=1}^{m_2} C_a(W, X) \otimes \tilde{e}_a,
\end{align*} 
where $C_a \in \lin{\R{p \times q} \otimes \R{m_2}, \R{n_1 \times \ell_1} \otimes \R{m_1}}{\R{\bar{n}_1 \times \bar{\ell}_1}}$ is a bilinear operator that defines the mechanics of the convolution. The specific form of $C_a$ is defined using (\ref{eqn:bigcrop}) and (\ref{eqn:Phi}) as follows:
\begin{align} \label{eqn:conv}
C_a(W, X) = \sum_{j=1}^{\bar{n}_1} \sum_{k=1}^{\bar{\ell}_1} \left\langle W_a, \Phi_{A_a}\left(\bigcrop_{1+(j-1)\Delta, 1+(k-1)\Delta}(X)\right)\right\rangle \bar{E}_{jk},
\end{align}
with $W = \sum_{a=1}^{m_2} W_a \otimes \tilde{e}_a$. The fixed vectors $\{A_a\}_{a=1}^{m_2}$, where $A_a \in \R{m_1}$ for each $a$, define the action of $\Phi_{A_a}$ and thus the mixing of feature maps. The choice of $\Delta$ defines the \emph{stride} of the convolution. 

The adjoints of the operators $(C \righthook X)$, $(W \lefthook C)$, and $(W \lefthook C_a)$ will be used in gradient calculations. The following theorems describe how to calculate them:
\begin{theorem}
Let $Y = \sum_{a=1}^{m_2} Y_a \otimes \tilde{e}_a \in \R{\bar{n}_1 \times \bar{\ell}_1} \otimes \R{m_2}$ and $X = \sum_{i=1}^{m_1} X_i \otimes e_i \in \R{n_1 \times \ell_1} \otimes \R{m_1}$. Then, 
\begin{align*}
(C \righthook X)^* \cdot Y = \sum_{a=1}^{m_2} \left\{\sum_{j=1}^{\bar{n}_1} \sum_{k=1}^{\bar{\ell}_1} \ip{Y_a}{\bar{E}_{jk}} \Phi_{A_a}(\bigcrop_{1+(j-1)\Delta, 1+(k-1)\Delta}(X))\right\} \otimes \tilde{e}_a.
\end{align*}
\begin{proof} Let $U = \sum_{a=1}^{m_2} U_a \otimes \tilde{e}_a \in \R{p \times q} \otimes \R{m_2}$. Then, 
\begin{align*}
\ip{Y}{(C \righthook X) \cdot U} &= \ip{Y}{C(U, X)} \\
&= \sum_{a=1}^{m_2} \ip{Y_a}{C_a(U, X)} \\
&= \sum_{a=1}^{m_2} \left\langle Y_a, \sum_{j=1}^{\bar{n}_1} \sum_{k=1}^{\bar{\ell}_1} \ip{U_a}{\Phi_{A_a} (\bigcrop_{1 + (j-1) \Delta, 1 + (k-1) \Delta}(X))} \bar{E}_{jk} \right\rangle \\
&= \sum_{a=1}^{m_2} \sum_{j=1}^{\bar{n}_1} \sum_{k=1}^{\bar{\ell}_1} \ip{Y_a}{\bar{E}_{jk}} \ip{\Phi_{A_a} (\bigcrop_{1 + (j-1) \Delta, 1 + (k-1) \Delta}(X))}{U_a} \\
&= \sum_{a=1}^{m_2} \left\langle \sum_{j=1}^{\bar{n}_1} \sum_{k=1}^{\bar{\ell}_1} \ip{Y_a}{\bar{E}_{jk}} \Phi_{A_a}(\bigcrop_{1 + (j-1) \Delta, 1 + (k-1) \Delta}(X)), U_a\right\rangle \\
&= \left\langle \sum_{a=1}^{m_2} \left\{\sum_{j=1}^{\bar{n}_1} \sum_{k=1}^{\bar{\ell}_1} \ip{Y_a}{\bar{E}_{jk}} \Phi_{A_a}(\bigcrop_{1 + (j-1) \Delta, 1 + (k-1) \Delta}(X))\right\} \otimes \tilde{e}_a, U\right\rangle.
\end{align*}
Since this is true for any $U$, the proof is complete.
\end{proof}
\end{theorem}

\begin{theorem} \label{thm:W_lh_C} Let $W = \sum_{a=1}^{m_2} W_a \otimes \tilde{e}_a \in \R{p \times q} \otimes \R{m_2}$ and $Y \in \R{\bar{n}_1 \times \bar{\ell}_1}$. Then, 
\begin{align*}
(W \lefthook C_a)^* \cdot Y = \sum_{j=1}^{\bar{n}_1} \sum_{k=1}^{\bar{\ell}_1} \ip{Y}{\bar{E}_{jk}} \bigcrop^*_{1+(j-1)\Delta, 1+(k-1)\Delta} \cdot \Phi_{A_a}^* \cdot W_a.
\end{align*}
Furthermore, for any $Z = \sum_{a=1}^{m_2} Z_a \otimes \tilde{e}_a \in \R{\bar{n}_1 \times \bar{\ell}_1} \otimes \R{m_2}$, 
\begin{align*}
(W \lefthook C)^* \cdot Z = \sum_{a=1}^{m_2} (W \lefthook C_a)^* \cdot Z_a.
\end{align*}
\begin{proof} Let $X = \sum_{i=1}^{m_1} X_i \otimes e_i \in \R{n_1 \times \ell_1} \otimes \R{m_1}$. Then,
\begin{align*}
\ip{Y}{(W \lefthook C_a) \cdot X} &= \ip{Y}{C_a(W, X)} \\
&= \sum_{j=1}^{\bar{n}_1} \sum_{k=1}^{\bar{\ell}_1} \ip{W_a}{\Phi_{A_a} \cdot \bigcrop_{1 + (j-1)\Delta, 1+ (r-1)\Delta} (X)}\ip{Y}{\bar{E}_{jk}} \\
&= \sum_{j=1}^{\bar{n}_1} \sum_{k=1}^{\bar{\ell}_1} \left\langle \ip{Y}{\bar{E}_{jk}} \bigcrop^*_{1 + (j-1)\Delta, 1 + (k-1) \Delta} \cdot \Phi^*_{A_a} \cdot W_a, X\right\rangle.
\end{align*}
Also,
\begin{align*}
\ip{Z}{(W \lefthook C) \cdot X} &= \ip{Z}{C(W, X)} \\
&= \sum_{a=1}^{m_2} \ip{Z_a}{C_a(W, X)} \\
&= \sum_{a=1}^{m_2} \ip{(W \lefthook C_a)^* \cdot Z_a}{X} \\
&= \left\langle \sum_{a=1}^{m_2} (W \lefthook C_a)^* \cdot Z_a, X\right\rangle.
\end{align*}
Both of the above results are true for a generic $X \in \R{n_1 \times \ell_1} \otimes \R{m_1}$, which completes the proof.
\end{proof}
\end{theorem}

\subsubsection{Elementwise Nonlinearity}
The $\bignl$ operator in (\ref{eqn:single_layer}) is an elementwise nonlinear function, $\bignl : \R{\bar{n}_1 \times \bar{\ell}_1} \otimes \R{m_2} \rightarrow \R{\bar{n}_1 \times \bar{\ell}_1} \otimes \R{m_2}$, that operates as follows:
\begin{align} \label{eqn:bignl}
\bignl\left(\sum_{a=1}^{m_2} Y_a \otimes \tilde{e}_a\right) = \sum_{a=1}^{m_2} \smallnl(Y_a) \otimes \tilde{e}_a, 
\end{align}
where $\smallnl : \R{\bar{n}_1 \times \bar{\ell}_1} \rightarrow \R{\bar{n}_1 \times \bar{\ell}_1}$ is some elementwise nonlinear function, which can be written as $\smallnl(Y_a) = \sum_{j=1}^{\bar{n}_1} \sum_{k=1}^{\bar{\ell}_1} \bar{\smallnl}(\ip{Y_a}{\bar{E}_{jk}}) \bar{E}_{jk}$. The map $\bar{\smallnl} : \R{} \rightarrow \R{}$ defines the nonlinear action. Common choices for $\bar{\smallnl}$ include the ramp function $\max(0,x)$ (also known as the \emph{rectifier}), the sigmoidal function, or hyperbolic tangent, for example. 

Some more maps are defined to assist in the calculation of the derivative $\D \bignl$ of $\bignl$. The elementwise \emph{first} and \emph{second} derivatives, $\bignl'$ and $\bignl''$, are maps of the same dimension as $\bignl$, defined with $\bar{\smallnl}$ replaced by $\bar{\smallnl}'$ and $\bar{\smallnl}''$ in the above formulation, respectively. Furthermore, consider a bilinear map $\odot \in \lin{\R{\bar{n}_1 \times \bar{\ell}_1} \otimes \R{m_2}, \R{\bar{n}_1 \times \bar{\ell}_1} \otimes \R{m_2}}{\R{\bar{n}_1 \times \bar{\ell}_1} \otimes \R{m_2}}$ that operates on $v = \sum_{a=1}^{m_2} v_a \otimes \tilde{e}_a$ and $w = \sum_{a=1}^{m_2} w_a \otimes \tilde{e}_a$ --- both in $\R{\bar{n}_1 \times \bar{\ell}_1} \otimes \R{m_2}$ --- according to:
\[
v \odot w = \sum_{a=1}^{m_2} \left(\sum_{j=1}^{\bar{n}_1} \sum_{k=1}^{\bar{\ell}_1} \ip{v_a}{\bar{E}_{jk}} \ip{w_a}{\bar{E}_{jk}} \bar{E}_{jk} \right) \otimes e_a.
\]
This is an extension of the Hadamard product to the tensor product space. The map $\D\bignl$ and its adjoint are now easy to calculate.
\begin{theorem}
For any $v$ and $z \in \R{\bar{n}_1 \times \bar{\ell}_1} \otimes \R{m_2}$,
\[
\D\bignl(z) \cdot v = \bignl'(z) \odot v.
\]
Furthermore, $\D\bignl(z)$ is self-adjoint, i.e. $\D^*\bignl (z)= \D\bignl (z)$. 
\begin{proof}
Let $z = \sum_{a=1}^{m_2} z_a \otimes \tilde{e}_a$ and $v = \sum_{a=1}^{m_2} v_a \otimes \tilde{e}_a$, where $z_a, v_a \in \R{\bar{n}_1 \times \bar{\ell}_1}$ for each $a$. Then,
\begin{align*}
\D\bignl(z) \cdot v &= \dd{}{t} \bignl (z + tv) \eval{t=0} \\
&= \dd{}{t} \sum_{a=1}^{m_2} \left(\sum_{j=1}^{\bar{n}_1} \sum_{k=1}^{\bar{\ell}_1}\bar{\smallnl}(\ip{z_a + t v_a}{\bar{E}_{jk}}) \bar{E}_{jk}\right) \otimes \tilde{e}_a \eval{t=0} \\
&= \sum_{a=1}^{m_2} \left(\sum_{j=1}^{\bar{n}_1} \sum_{k=1}^{\bar{\ell}_1} \bar{\smallnl}' (\ip{z_a}{\bar{E}_{jk}}) \ip{v_a}{\bar{E}_{jk}} \bar{E}_{jk} \right) \otimes \tilde{e}_a \\
&= \bignl'(z) \odot v,
\end{align*}
where the final line follows from the definition of the Hadamard product and the elementwise first derivative $\bignl'(z)$. To prove that $\D\bignl(z)$ is self-adjoint, first note that it is not hard to show that $\ip{y}{v \odot w} = \ip{v \odot y}{w}$, for any $v, w,$ and $y$ in the same space. Thus, for any $y \in \R{\bar{n}_1 \times \bar{\ell}_1} \otimes \R{m_2}$, 
\begin{align*}
\ip{y}{\D\bignl(z) \cdot v} &= \ip{y}{\bignl'(z) \odot v} \\
&= \ip{\bignl'(z) \odot y}{v} \\
&= \ip{\D\bignl(z) \cdot y}{v}.
\end{align*}
This proves that $\D^*\bignl(z) = \D\bignl(z)$. 
\end{proof}
\end{theorem}  

\subsubsection{Pooling Operator}
The $\bigpool$ operator in (\ref{eqn:single_layer}) is known as the \emph{pooling} operator, and its purpose is to reduce the size of the feature maps at each layer. Only \emph{linear} pooling is considered in this paper (the framework does extend to the nonlinear case though), so that $\bigpool \in \lin{\R{\bar{n}_1 \times \bar{\ell}_1} \otimes \R{m_2}}{\R{n_2 \times \ell_2} \otimes \R{m_2}}$ operates as:
\begin{align} \label{eqn:bigpool}
\bigpool\left(\sum_{a=1}^{m_2} Y_a \otimes \tilde{e}_a\right) = \sum_{a=1}^{m_2} \smallpool(Y_a) \otimes \tilde{e}_a
\end{align}
for $\sum_{a=1}^{m_2} Y_a \otimes \tilde e_a \in \R{\bar n_1 \times \bar \ell_1} \otimes \R{m_2}$. Here $\smallpool \in \lin{\R{\bar{n}_1 \times \bar{\ell}_1}}{\R{n_2 \times \ell_2}}$ operates in the same way for each feature map $Y_a$. The operator $\smallpool$ acts on disjoint $r \times r$ neighbourhoods that form a partition of the input $Y_a$, with one output from each neighbourhood. This implies that $\bar{n}_1 = r n_2$ and $\bar{\ell}_1 = r \ell_2$ (assuming that $r | \bar{n}_1$ and $r | \bar{\ell}_1$).

One type of linear pooling is \emph{average pooling}, which involves taking the average over all elements in the $r \times r$ neighbourhoods. This can be represented using (\ref{eqn:smallcrop}) as: 
\begin{align} \label{eqn:smallpool}
\smallpool(Y_a) = \frac{1}{r^2} \sum_{j=1}^{n_2} \sum_{k=1}^{\ell_2} \ip{\mathbf{1}_r}{ \smallcrop_{1+(j-1)r, 1+(k-1)r}(Y_a)} \hat{E}_{jk},
\end{align}
 where the operator $\smallcrop_{jk} \in \lin{\R{\bar{n}_1 \times \bar{\ell}_1}}{\R{r \times r}}$ is defined in (\ref{eqn:smallcrop}) with $p = q = r$ and 
 \[
 \mathbf{1}_r = \sum_{j = 1}^r  \sum_{k = 1}^r \bar{E}_{jk} \in \R{r \times r}.
 \]
  If $\{\bar{E}_{jk}\}_{jk}$ is the standard basis, $\mathbf{1}_r$ is the all-ones matrix. 

The adjoint $\Psi^*$ of the average pooling operator $\Psi$ can be computed using the following theorem.
\begin{theorem} \label{thm:bigpool}
Let $Z = \sum_{a=1}^{m_2} Z_a \otimes \tilde{e}_a \in \R{n_2 \times \ell_2} \otimes \R{m_2}$. Then, using \emph{(\ref{eqn:embed})} with $\Em_{c,d} : \R{r \times r} \rightarrow \R{\bar{n}_1 \times \bar{\ell_1}}$, 
\begin{align*}
\bigpool^* \cdot Z = \frac{1}{r^2} \sum_{a=1}^{m_2} \left\{\sum_{j=1}^{n_2} \sum_{k=1}^{\ell_2} \ip{Z_a}{\hat{E}_{jk}} \Em_{1+(j-1)r, 1+(k-1)r}(\mathbf{1}_r)\right\} \otimes \tilde{e}_a.
\end{align*} 
\begin{proof}
First, let $\gamma_{jkr} = (1 + (j-1)r, 1 + (k-1)r)$ for notational convenience. Then, for any $Y = \sum_{a=1}^{m_2} Y_a \otimes \tilde{e}_a \in \R{\bar{n}_1 \times \bar{\ell}_1} \otimes \R{m_2}$, 
\begin{align*}
\ip{Z}{\bigpool \cdot Y} &=\sum_{a=1}^{m_2} \left\langle Z_a, \frac{1}{r^2} \sum_{j=1}^{n_2} \sum_{k=1}^{\ell_2} \ip{\mathbf{1}_r}{\smallcrop_{\gamma_{jkr}} (Y_a)} \hat{E}_{jk}\right\rangle \\
&= \frac{1}{r^2} \sum_{a=1}^{m_2} \sum_{j=1}^{n_2} \sum_{k=1}^{\ell_2} \ip{Z_a}{\hat{E}_{jk}} \ip{\mathbf{1}_r}{\smallcrop_{\gamma_{jkr}}(Y_a)} \\
&= \frac{1}{r^2} \sum_{a=1}^{m_2} \left\langle \sum_{j=1}^{n_2} \sum_{k=1}^{\ell_2} \ip{Z_a}{\hat{E}_{jk}} \smallcrop^*_{\gamma_{jkr}}(\mathbf{1}_r), Y_a\right\rangle \\
&= \left\langle \frac{1}{r^2} \sum_{a=1}^{m_2} \left\{\sum_{j=1}^{n_2} \sum_{k=1}^{\ell_2} \ip{Z_a}{\hat{E}_{jk}} \Em_{\gamma_{jkr}}(\mathbf{1}_r)\right\} \otimes \tilde{e}_a, Y \right\rangle.
\end{align*}
Since this is true for any $Y$, the proof is complete. 
\end{proof}
\end{theorem}

\subsubsection{Single-Layer Derivatives}
The derivatives of a generic layer $f(X;W,B)$, as described in (\ref{eqn:single_layer}), with respect to $X$, $W$, and $B$ are presented in the following theorem.
\begin{theorem} \label{thm:Df}
\hspace{0.1mm}
\begin{enumerate}
\item $\D f(X;W,B) = \bigpool \cdot \D\bignl(C(W,X) + B) \cdot (W \lefthook C).$
\item $\nabla_Wf(X;W,B) = \bigpool \cdot \D\bignl(C(W, X) + B) \cdot (C \righthook X).$
\item $\nabla_Bf(X;W, B) = \bigpool \cdot \D\bignl(C(W, X) + B).$
\end{enumerate}
\begin{proof}
These are all direct consequences of the chain rule and linearity of the derivative for the function $f$ given in \eqref{eqn:single_layer}. 
\end{proof}
\end{theorem}

The adjoints of the above operators can be calculated using the reversing property of the adjoint operator $*$. 

\clearpage

\begin{theorem} \label{thm:Df:star}
\hspace{0.1mm}
\begin{enumerate}
\item $\D^* f(X;W,B) = (W \lefthook C)^* \cdot \D\bignl(C(W,X) + B) \cdot \bigpool^*.$
\item $\nabla_W^*f(X;W,B) = (C \righthook X)^* \cdot  \D\bignl(C(W, X) + B) \cdot \bigpool^*.$
\item $\nabla_B^* f(X;W, B) =  \D\bignl(C(W, X) + B) \cdot \bigpool^* .$
\end{enumerate}
\end{theorem} 

\subsection{Multiple Layers}
Suppose now that the network consists of $L$ layers. Denote the actions of the $t^{th}$ layer as $X^{t+1} = f_t(X^t)$, where $X^t \in \R{n_t \times \ell_t} \otimes \R{m_t}$ and $X^1$ is one point in the input data. The layer map $f_t : \R{n_t \times \ell_t} \otimes \R{m_t} \rightarrow \R{n_{t+1} \times \ell_{t+1}} \otimes \R{m_{t+1}}$ can be given explicitly as:
\begin{align} \label{eqn:f_t}
f_t(X^t) \coloneqq \bigpool_t\left(\bignl_t(C^t(W^t, X^t) + B^t)\right).
\end{align}
Here, $W^t \in \R{p_t \times q_t} \otimes \R{m_{t+1}}$ and $B^t \in \R{\bar{n}_t \times \bar{\ell}_t} \otimes \R{m_{t+1}}$. Note that the pooling operator $\bigpool_t$, the nonlinearity $\bignl_t$, and the convolution operator $C^t$ are layer-dependent. The entire network's actions can be denoted as:
\[
F(X; \theta) \coloneqq f_L \circ \cdots \circ f_1(X),
\] 
where $\theta \coloneqq \{W^1, \ldots, W^L, B^1, \ldots, B^L\}$ is the parameter set and $X \equiv X^1$ is the input data. 

\subsubsection{Final Layer}
Classification is often the goal of a CNN, thus assume that there are $N$ classes. This implies the following: $m_{L+1} = N$,  $\bar{n}_L = \bar{\ell}_L = n_{L+1} = \ell_{L+1} = 1$, and $F(X; \theta) \in \R{N}$. The final layer is assumed to be  fully connected, which aligns with the form given in (\ref{eqn:f_t}) if the cropping operator (\ref{eqn:bigcrop}) and pooling operator (\ref{eqn:bigpool}) for the final layer --- $\bigcrop^L_{jk}$ and $\bigpool_L$, respectively --- are identity maps. Also, $A_a^L \in \R{m_L}$ defining the mixing operator $\Phi^L_{A^L_a}$ in (\ref{eqn:conv}) is $A^L_a = \sum_{i=1}^{m_L} e_i^L$ for each $a$. Then, the final layer is given as:
\[
f_L(X^L) = \bignl_L(C^L(W^L, X^L) + B^L) \equiv \sum_{a=1}^{N} \smallnl_L\left(C_a^L(W^L, X^L) + B_a^L\right) e_a^{L+1}, 
\]
where $\{e_a^{L+1}\}_a$ is a basis for $\R{N}$, and $C_a^L(W^L, X^L) = \sum_{i=1}^{m_L} \ip{W_a^L}{X_i^L}$. Note that $\smallnl_L : \R{} \rightarrow \R{}$. It is also important to note that this shows that simpler, fully-connected neural networks are just a special case of convolutional neural networks.

\subsection{Loss Function \& Backpropagation}
While training a CNN, the goal is to optimize some loss function $J$ with respect to the parameters $\theta$. For example, consider 
\[
J(X; \theta) \coloneqq \frac{1}{2} \norm{y - F(X; \theta)}^2 = \frac{1}{2} \ip{y - F(X; \theta)}{y - F(X; \theta)},
\]
where $y$ represents the given data and $F(X;\theta)$ is the prediction. Gradient descent is used to optimize the loss function, so it is important to calculate the gradient of $J$ with respect to each of the parameters. For this, define maps $\omega_t$ and $\head_t$ as:
\begin{align}
\omega_t \coloneqq f_L \circ \cdots \circ f_t, \qquad \head_t \coloneqq f_t \circ \cdots \circ f_1 \label{eqn:omega_phi}
\end{align}
for $t= 1, \ldots, L$,
which satisfy (\ref{eqn:recursive_rel}). Assume $\omega_{L+1}$ and $\head_0$ are identity maps for the sake of convenience. Then, for any $U^t \in \R{n_t \times \ell_t} \otimes \R{m_t}$,
\begin{align*}
\nabla_{W^t} J(X; \theta) \cdot U^t = \ip{F(X; \theta) - y}{\nabla_{W^t} F(X; \theta) \cdot U^t} =  \ip{\nabla_{W^t}^*F(X;\theta) \cdot (F(X; \theta) - y)}{U^t}.
\end{align*}
Since this holds for any $U^t$, 
\begin{align} \label{eqn:dJdW}
\nabla_{W^t} J(X; \theta) = \nabla_{W^t}^*F(X;\theta) \cdot (F(X; \theta) - y)
\end{align}
by the same logic used to derive (\ref{eqn:nabla_J}) from (\ref{eqn:nabla_J_u}).
Differentiating  $F(X; \theta) = \omega_{t+1} \circ f_t \circ \head_{t-1}(X)$ with respect to $W^t$ produces
\begin{align} \label{eqn:dFdW}
\nabla_{W^t} F(X; \theta) = \D\omega_{t+1}(X^{t+1}) \cdot \nabla_{W^t}f_t(X^t),
\end{align}
where $X^t = \head_{t-1}(X)$ and $X^{t+1} = f_t(X^t) = \head_t(X)$.  Taking the adjoint of (\ref{eqn:dFdW}) yields
\begin{equation}\label{eqn:dFdW:adjoint}
\nabla_{W^t}^* F(X; \theta) = \nabla_{W^t}^*f_t(X^t) \cdot  \D^*\omega_{t+1}(X^{t+1}), 
\end{equation}
which can be substituted into (\ref{eqn:dJdW}). Then, the final step in computing (\ref{eqn:dJdW}) involves computing $\D^*\omega_{t+1}$ in (\ref{eqn:dFdW:adjoint}), which can be done recursively:
\begin{equation}\label{Domega:adjoint:recursion}
\D^*\omega_t(X^t) = \D^*f_t(X^t) \cdot \D^*\omega_{t+1}(X^{t+1}).
\end{equation}
This  comes from taking the derivative and then the adjoint of the relationship $\omega_{t} = \omega_{t+1} \circ f_t$. Note that 
 $\nabla_{W^t}^*f_t (X^t)$ and $\D^*f_t(X^t)$ in (\ref{eqn:dFdW:adjoint}) and (\ref{Domega:adjoint:recursion}) are calculated using Theorem \ref{thm:Df:star}. Since $\nabla_{W^t}J(X; \theta)$ can be calculated, gradient descent can be performed. One iteration of a gradient descent algorithm to update $B^t$ and $W^t$ for all $t \in \{1, \ldots, L\}$ is given in Algorithm \ref{alg:CNN_first_grad_desc}. The method for calculating $\nabla_{B^t}J(X; \theta)$ is not explicitly shown in the derivation, but is a simpler version of $\nabla_{W^t}J(X;\theta)$ and is included in the algorithm. The algorithm can be extended to a batch of points by summing the contribution to $\nabla J$ from each input point $X$. Note that $\eta \in \R{}$ is the learning rate. 
 
\begin{algorithm}
\caption{One iteration of gradient descent for a CNN}
\label{alg:CNN_first_grad_desc}
\begin{algorithmic}
\Function{Descent Iteration}{$X, y, W^1, \ldots, W^L, B^1, \ldots, B^L, \eta$}
\State $X^1 \gets X$
\For {$t \in \{1, \ldots, L\}$} \Comment $X^{L+1} = F(X; \theta)$
\State $Z^t \gets C^t(W^t, X^t) + B^t$
\State $X^{t+1} \gets \bigpool_t \left(\bignl_t(Z^t)\right)$ \Comment $f_t$ from \eqref{eqn:f_t}
\EndFor
\For {$t \in \{L, \ldots, 1\}$}
\State $\tilde{W}^t \gets W^t$ \Comment Store old $W^t$ for updating $W^{t-1}$
\If {$t = L$} \Comment $e = \D^*\tail_{t+1}(X^{t+1}) \cdot \left(X^{L+1} - y\right)$
\State $e \gets x^{L+1} - y$   \Comment $\tail_{L+1} = \textup{identity}$ 
\Else
\State $e \gets \left(\tilde{W}^{t+1} \lefthook C^{t+1}\right)^* \cdot \left(\bignl'_{t+1}(Z^{t+1}) \odot \left(\bigpool_{t+1}^* \cdot e\right)\right)$ \Comment \eqref{Domega:adjoint:recursion} \& Thm \ref{thm:Df:star}, update with $\tilde{W}^{t+1}$  
\EndIf
\State $\nabla_{B^t} J(X; \theta) \gets \left(\bignl_t'(Z^t) \odot (\bigpool_t^*\cdot e)\right)$
\State $\nabla_{W^t} J(X; \theta) \gets \left(C^t \righthook X^t\right)^* \cdot \left(\bignl_t'(Z^t) \odot (\bigpool_t^* \cdot e) \right)$ \Comment \eqref{eqn:dFdW:adjoint} \& Thm \ref{thm:Df:star}
\State $B^t \gets B^t - \eta \nabla_{B^t} J(X; \theta)$
\State $W^t \gets W^t - \eta \nabla_{W^t} J(X; \theta)$
\EndFor
\EndFunction
\end{algorithmic}
\end{algorithm}

\subsection{Higher-Order Loss Functions}
Suppose that another term is added to the loss function to penalize the first-order derivative of $F(X;\theta)$, as in \cite{rifai2011manifold} or \cite{simard1992tangent} for example. This can be represented using 
\[
R(X; \theta) \coloneqq \frac{1}{2} \norm{\D F(X; \theta) \cdot V_X - \beta_X}^2,
\] 
for some $V_X \in \R{n_1 \times \ell_1} \otimes \R{m_1}$ and $\beta_X \in \R{N}$. When $\beta_X = 0$, minimizing $R(X, \theta)$ promotes invariance of the network in the direction of $V_X$. This can be useful in image classification, for example, where the class of image is expected to be invariant with respect to rotation. In this case, $V_X$ would be an infinitesimal generator of rotation. This new term $R$ can be added to $J$ to create a new loss function 
\begin{align} \label{eqn:curly_J}
\mathcal{J} \coloneqq J + \lambda R,
\end{align}
where $\lambda \in \R{}$ determines the amount that the higher-order term contributes to the loss function. Note that $R$ could be extended to contain multiple terms as:
\begin{align} \label{eqn:R_mult_terms}
R(X;\theta) = \sum_{(V_X, \beta_X) \in \mathcal B_X} \frac{1}{2} \norm{\D F(X; \theta) \cdot V_X - \beta_X}^2,
\end{align}
where $\mathcal B_X$ is a finite set of pairs $(V_X, \beta_X)$ for each $X$.

The gradient of $R$ with respect to the parameters must now be taken. This can calculated for a generic parameter $\theta^t$, which is one of $W^t$ or $B^t$:
\begin{align*}
\nabla_{\theta^t}R(X; \theta) \cdot U^t = \ip{\D F(X; \theta) \cdot V_X - \beta_X}{\left(\nabla_{\theta^t} \D F(X; \theta) \righthook V_X\right) \cdot U^t}, 
\end{align*}
for all $U^t$ in the same space as $\theta^t$. Again, in the same way that \eqref{eqn:nabla_J} was derived from \eqref{eqn:nabla_J_u}, 
\begin{align}  \label{eqn:dRdW}
\nabla_{\theta^t}R(X; \theta) = \left(\nabla_{\theta^t} \D F(X; \theta) \righthook V_X\right)^* \cdot \left(\D F(X; \theta) \cdot V_X - \beta_X\right).
\end{align}  
Before \eqref{eqn:dRdW} can be computed, however, some preliminary results will be given. 

\begin{theorem} \label{thm:D_nab_f} Let $f$ be defined as in \eqref{eqn:single_layer}, and $V \in \R{n_1 \times \ell_1} \otimes \R{m_1}$. Let $Z = C(W, X) + B$. Then,
\begin{align}
(V \lefthook \D \nabla_{W}f(X; W, B)) &= \bigpool \cdot \left(C(W,V) \lefthook \D^2S(Z)\right) \cdot (C \righthook X) + \bigpool \cdot \D S(Z) \cdot \left(C \righthook V\right), \label{eqn:V_lh_D_W} \\
(V \lefthook \D \nabla_B f(X; W, B)) &= \bigpool \cdot \left( C(W, V) \lefthook \D^2 \bignl(Z) \right), \label{eqn:V_lh_D_B} \\
(V \lefthook \D^2 f(X; W, B)) &= \bigpool \cdot \left(C(W, V) \lefthook \D^2S(Z)\right) \cdot (W \lefthook C). \label{eqn:D2f}
\end{align}
\begin{proof}
Let $U \in \R{p \times q} \otimes \R{m_2}$. Then, prove \eqref{eqn:V_lh_D_W} directly:
\begin{align*}
\left(V \lefthook \D\nabla_W f(X; W, B)\right)\cdot U &= \D \left(\nabla_W f(X; W, B) \cdot U\right) \cdot V \\
&= \D \left[\bigpool \cdot \D\bignl (C(W, X) + B) \cdot C(U, X)\right] \cdot V \\
&= \bigpool \cdot \D^2\bignl(Z) \cdot (C(W, V), C(U, X)) + \bigpool \cdot \D\bignl(Z) \cdot C(U, V) \\
&= \bigpool \cdot \left[ \left(C(W,V) \lefthook \D^2 \bignl(Z)\right) \cdot (C \righthook X) + \D\bignl(Z) \cdot (C \righthook V) \right] \cdot U.
\end{align*}
This is true for any $U$, so equation \eqref{eqn:V_lh_D_W} is proven. Equation \eqref{eqn:V_lh_D_B} can be proven similarly, so its proof is omitted. Also, let $\tilde{V} \in \R{n_1 \times \ell_1} \otimes \R{m_1}$. Then, equation \eqref{eqn:D2f} can also be proven directly:
\begin{align*}
\left(V \lefthook \D^2 f(X; W, B)\right) \cdot \tilde{V} &= \D \left( \bigpool \cdot \D\bignl(C(W, X) + B) \cdot C(W, \tilde{V}) \right) \cdot V \\
&= \bigpool \cdot \D^2 \bignl(Z) \cdot (C(W, V), C(W, \tilde{V})) \\
&= \bigpool \cdot \left( C(W, V) \lefthook \D^2 \bignl(Z) \right) \cdot \left(W \lefthook C\right) \cdot \tilde{V}.
\end{align*}
This is true for any $\tilde{V}$, so the proof is completed. 
\end{proof}
\end{theorem}
The next lemma shows how to actually calculate $\D^2\bignl(Z)$ so that the above equations can be computed. 

\begin{lemma} \label{lem:D2S}
For any $X, V$ and $\tilde{V} \in \R{\bar{n}_1 \times \bar{\ell}_1} \otimes \R{m_2}$ with $\bignl$ defined in \eqref{eqn:bignl},
\[
\D^2\bignl(X) \cdot (V, \tilde{V}) = \bignl''(X) \odot V \odot \tilde{V},
\]
where $\bignl''$ is defined similarly to $\bignl$, but with $\bar{\smallnl}''$ replacing $\bar{\smallnl}$. Furthermore, $(V \lefthook \D^2S(X))$ is self-adjoint, i.e. $(V \lefthook \D^2S(X))^* = (V \lefthook \D^2S(X))$. 
\begin{proof}
From the definition of the second derivative, 
\begin{align*}
\D^2\bignl(X) \cdot (V, \tilde{V}) &= \D \left(\D\bignl(X) \cdot \tilde{V} \right) \cdot V \\
&= \D\left(\bignl'(X) \odot \tilde{V}\right) \cdot V \\
&= \left(\bignl''(X) \odot \tilde{V} \right) \odot V,
\end{align*}
where the last equality follows from viewing $\bignl'(X) \odot \tilde{V}$ as an elementwise function in $X$. As for the adjoint, let $Y \in \R{\bar{n}_1 \times \bar{\ell}_1} \otimes \R{m_2}$. Then,
\begin{align*}
\ip{Y}{\left(V \lefthook \D^2 \bignl(X)\right) \cdot \tilde{V}} &= \ip{Y}{\bignl''(X) \odot V \odot \tilde{V}} \\
&= \ip{\bignl''(X) \odot V \odot Y}{\tilde{V}} \\
&= \ip{\left(V \lefthook \D^2 \bignl(X) \right) \cdot Y}{\tilde{V}}.
\end{align*}
This proves that $\left(V \lefthook \D^2 \bignl(X) \right)$ is self-adjoint.
\end{proof}
\end{lemma}

The adjoints of the  equations in Theorem \ref{thm:D_nab_f} can now easily be calculated using the above lemma and the reversing property of the adjoint operator. 
\begin{theorem}\label{thm:D_nab_f:star}
Let $f$ be defined as in \eqref{eqn:single_layer}, and $V \in \R{n_1 \times \ell_1} \otimes \R{m_1}$. Let $Z = C(W, X) + B$. Then,
\begin{align*}
(V \lefthook \D \nabla_{W}f(X; W, B))^* &= (C \righthook X)^* \cdot \left(C(W,V) \lefthook \D^2S(Z)\right) \cdot \bigpool^*+ \left(C \righthook V\right)^* \cdot \D S(Z) \cdot \bigpool^*,  \\
(V \lefthook \D \nabla_B f(X; W, B))^* &=  \left( C(W, V) \lefthook \D^2 \bignl(Z) \right) \cdot \bigpool^*,  \\
(V \lefthook \D^2 f(X; W, B))^* &=  (W \lefthook C)^* \cdot \left(C(W, V) \lefthook \D^2S(Z)\right) \cdot \bigpool^*. 
\end{align*}
\end{theorem}

Now, propagation through the tangent network can be described in the spirit of \cite{simard1992tangent}. \emph{Forward} propagation through the network can be computed recursively, using $\head_t = f_t \circ \head_{t-1}$:
\begin{align} \label{eqn:D_phi}
\D \head_t(X) = \D f_t(X^t) \cdot \D \head_{t-1}(X),
\end{align}
for any $t \in \{1, \ldots, L\}$ and $X \in \R{n_1 \times \ell_1} \otimes \R{m_1}$. \emph{Backward} propagation through the tangent network is described in the next theorem.

\begin{theorem}\label{theorem:recursive:update:varphi:omega}
Let $f_t$ be defined as in \emph{(\ref{eqn:f_t})} and $\omega_t$ and $\head_t$ be defined as in \emph{(\ref{eqn:omega_phi})}. Then, for any $X, V \in \R{n_1 \times \ell_1} \otimes \R{m_1}$, and $t \in \{1, \ldots, L\}$, 
\begin{align}
\left((\D \head_{t-1}(X) \cdot V) \lefthook \D^2\omega_t(X^t) \right)^* &= \D^* f_t(X^t) \cdot \left( (\D\head_t(X) \cdot V) \lefthook \D^2\omega_{t+1}(X^{t+1}) \right)^* \nonumber \\
&\quad + \left( (\D\head_{t-1}(X) \cdot V) \lefthook \D^2f_t(X^t) \right)^* \cdot \D^* \omega_{t+1}(X^{t+1}), \label{eqn:tgt_backprop}
\end{align}
where $X^t = \head_{t-1}(X)$. Also, $\left( (\D\head_L(X) \cdot V) \lefthook \D^2\tail_{L+1}(X^{L+1})\right)^*$ is the zero operator. 
\begin{proof}
Since $\tail_{L+1}$ is the identity, its second derivative is the zero operator. Now consider the case when $t \in \{1, \ldots, L\}$. Take any $\tilde{X}, Y,$ and $\tilde{Y} \in \R{n_t \times \ell_t} \otimes \R{m_t}$. Then, 
\begin{align*}
\left(\tilde{Y} \lefthook \D^2\tail_t(\tilde{X}) \right) \cdot Y &= \D\left( \D\left(\tail_{t+1} \circ f_t\right) (\tilde{X}) \cdot Y  \right) \cdot \tilde{Y} \\
&= \D\left(\D\tail_{t+1}(f_t(\tilde{X})) \cdot \D f_t(\tilde{X}) \cdot Y \right) \cdot \tilde{Y} \\
&= \D^2 \tail_{t+1} (f_t(\tilde{X})) \cdot \left(\D f_t(\tilde{X}) \cdot \tilde{Y}, \D f_t(\tilde{X}) \cdot Y\right)  \\
&\qquad + \D\tail_{t+1}(f_t(\tilde{X})) \cdot \D^2 f_t(\tilde{X}) \cdot (\tilde{Y}, Y) \\
&= \left( \left(\D f_t(\tilde{X}) \cdot \tilde{Y}\right) \lefthook \D^2 \tail_{t+1}(f_t(\tilde{X})) \right) \cdot \D f_t(\tilde{X}) \cdot Y \\
& \qquad + \D\omega_{t+1}(f_t(\tilde{X})) \cdot \left(\tilde{Y} \lefthook \D^2 f_t(\tilde{X}) \right) \cdot Y,
\end{align*}
where the third equality follows from the product rule. Removing the trailing $Y$ from both sides, and setting $\tilde{Y} = \D\head_{t-1}(X) \cdot V$ and $\tilde{X} = \head_{t-1}(X) = X^t$, 
\begin{align*}
\left( (\D\head_{t-1}(X) \cdot V) \lefthook \D^2\tail_t(X^t) \right) &= \left( (\D\head_t(X) \cdot V) \lefthook \D^2\tail_{t+1}(X^{t+1}) \right) \cdot \D f_t(X^t) \\
&\qquad + \D\tail_{t+1}(X^{t+1}) \cdot \left( (\D\head_{t-1}(X)\cdot V) \lefthook \D^2 f_t(X^t) \right),
\end{align*}
since $\D\head_t(X) = \D f_t(X^t) \cdot \D\head_{t-1}(X)$ and $X^{t+1} = f_t(X^t) = \head_t(X)$. Taking the adjoint of this result completes the proof. 
\end{proof}
\end{theorem}

Note that calculating \eqref{eqn:tgt_backprop} involves taking the adjoint of \eqref{eqn:D2f}, which can be done using Theorem \ref{thm:D_nab_f:star}  along with Theorems \ref{thm:W_lh_C} and \ref{thm:bigpool} and Lemma \ref{lem:D2S}. The above results are crucial for the next theorem, which is the main result.
\begin{theorem} \label{thm:main} 
Suppose $V$ and $X \in \R{n_1 \times \ell_1} \otimes \R{m_1}$, $t \in \{1,\ldots, L\}$, and $F$, $\head_t$, and $\tail_t$ are defined as in \eqref{eqn:recursive_rel}. Then, for a generic parameter $\theta^t \in \{W^t, B^t\}$, 
\begin{align}
\left( \nabla_{\theta^t} \D F(X; \theta) \righthook V \right)^* &= \nabla^*_{\theta^t} f_t(X^t) \cdot \left( (\D \head_t(X) \cdot V) \lefthook \D^2\omega_{t+1}(X^{t+1}) \right)^* \nonumber \\
&\quad + \left( (\D\head_{t-1}(X) \cdot V) \lefthook \D\nabla_{\theta^t} f_t(X^t) \right)^* \cdot \D^*\omega_{t+1}(X^{t+1}), \label{eqn:nab_D_F}
\end{align} 
where $X^t = \head_{t-1}(X)$.
\begin{proof}
For any $U$ in the same space as $\theta^t$,
\begin{align*}
\left(\nabla_{\theta^t} \D F(X; \theta) \righthook V\right) \cdot U &= \D \left(\nabla_{\theta^t} F(X; \theta) \cdot U \right) \cdot V \\
&= \D \left( \D \tail_{t+1}(\head_t(X)) \cdot \nabla_{\theta^t} f_t(\head_{t-1}(X)) \cdot U\right) \cdot V \\
&= \D^2\tail_{t+1}(\head_t(X)) \cdot (\D\head_t(X) \cdot V, \nabla_{\theta^t} f_t(\head_{t-1}(X)) \cdot U) \\
&\qquad + \D\tail_{t+1}(\head_t(X)) \cdot \D\nabla_{\theta^t} f_t(\head_{t-1}(X)) \cdot (\D\head_{t-1}(X) \cdot V, U) \\
&= \left( (\D\head_t(X) \cdot V) \lefthook \D^2\tail_{t+1}(X^{t+1}) \right) \cdot \nabla_{\theta^t}f_t(X^t) \cdot U \\
&\qquad + \D\tail_{t+1}(X^{t+1}) \cdot \left( (\D\head_{t-1}(X)\cdot V) \lefthook \D\nabla_{\theta^t} f_t(X^t) \right) \cdot U,
\end{align*}
where the final equality follows since $X^t = \head_{t-1}(X)$ for all $t \in \{1,\ldots, L+1\}$. Removing the trailing $U$ from both sides and taking the adjoint produces equation \eqref{eqn:nab_D_F}. 
\end{proof}
\end{theorem}
Note that in Equation \eqref{eqn:nab_D_F}, $\nabla_{\theta_t}^*f_t$ and $\D\nabla_{\theta_t} f_t$ can be replaced by their corresponding expressions in Theorem \ref{thm:Df} and \ref{thm:D_nab_f}, respectively, once $\theta_t$ is replaced by one of $W_t$ or $B_t$. 
Then, \eqref{eqn:dRdW} can be computed with Theorem \ref{thm:main}, where $\D F(X; \theta) = \D\head_L(X)$ is computed recursively by \eqref{eqn:D_phi}. Algorithm \ref{alg:CNN_second_grad_desc} shows one iteration of a gradient descent algorithm to optimize $\mathcal{J}$ defined in \eqref{eqn:curly_J} for one point $X$. This algorithm extends to a batch of updates, and for $R$ defined with multiple $(V_X, \beta_X)$ pairs as in \eqref{eqn:R_mult_terms}. 

\begin{algorithm}
\caption{One iteration of gradient descent for a higher-order CNN}
\label{alg:CNN_second_grad_desc}
\begin{algorithmic}
\Function{Descent Iteration}{$X, y, V_X, \beta_X, W^1, \ldots, W^L, B^1, \ldots, B^L, \eta, \lambda$}
\State $X^1 \gets X$
\State $V^1 \gets V_X$ \Comment $V^t = \D\head_{t-1}(X) \cdot V_X$
\For {$t \in \{1, \ldots, L\}$} \Comment $X^{L+1} = F(X; \theta); V^{L+1} = \D F(X: \theta) \cdot V_X$
\State $Z^t \gets C^t(W^t, X^t) + B^t$
\State $X^{t+1} \gets \bigpool_t \left(\bignl_t(Z^t)\right)$ \Comment $f_t$ from \eqref{eqn:f_t}
\State $V^{t+1} \gets \bigpool_t \left( \bignl_t'(Z^t) \odot C^t(W^t, V^t)\right)$ 
\Comment \eqref{eqn:D_phi} with Thm. \ref{thm:Df} \EndFor
\For {$t \in \{L, \ldots, 1\}$}
\State $\tilde{W}^t \gets W^t$ \Comment Store old $W^t$ for updating $W^{t-1}$
\If {$t = L$} \Comment $\tail_{L+1} = \textup{identity}$ 
\State $e_y \gets x^{L+1} - y$   \Comment $e_y = \D^*\tail_{t+1}(X^{t+1}) \cdot \left(X^{L+1} - y\right)$
\State $e_w \gets 0$ \Comment $e_w = \left( V^{t+1} \lefthook \D^2\tail_{t+1}(X^{t+1})\right)^* \cdot (V^{L+1} - \beta_X)$
\State $e_v \gets V^{L+1} - \beta_X$ \Comment $e_v = \D^*\tail_{t+1}(X^{t+1}) \cdot (V^{L+1} - \beta_X)$
\State
\Else \Comment Update these with $\tilde{W}^{t+1}$
\State $e_y \gets \left(\tilde{W}^{t+1} \lefthook C^{t+1}\right)^* \cdot \left(\bignl'_{t+1}(Z^{t+1}) \odot \left(\bigpool_{t+1}^* \cdot e_y\right)\right)$ \Comment \eqref{Domega:adjoint:recursion} with Thm. \ref{thm:Df:star}
\State $e_w \gets \left(\tilde{W}^{t+1} \lefthook C^{t+1}\right)^* \cdot \left(\bignl'_{t+1}(Z^{t+1}) \odot \left(\bigpool_{t+1}^* \cdot e_w\right)\right)$ 
\State $\qquad \qquad + \left(\tilde{W}^{t+1} \lefthook C^{t+1}\right)^* \cdot \left(\bignl_{t+1}''(Z^{t+1}) \odot C^{t+1}(\tilde{W}^{t+1}, V^{t+1}) \odot \left(\bigpool^*_{t+1} \cdot e_v\right) \right)$
\State \Comment \eqref{eqn:tgt_backprop} with Thms. \ref{thm:Df:star} \& \ref{thm:D_nab_f:star}, use old $e_v$ to update 
\State 
\State $e_v \gets \left(\tilde{W}^{t+1} \lefthook C^{t+1}\right)^* \cdot \left(\bignl'_{t+1}(Z^{t+1}) \odot \left(\bigpool_{t+1}^* \cdot e_v\right)\right)$ \Comment \eqref{Domega:adjoint:recursion} with Thm. \ref{thm:Df:star}
\EndIf
\State $\nabla_{B^t} J(X; \theta) \gets \left(\bignl_t'(Z^t) \odot (\bigpool_t^*\cdot e_y)\right)$
\State $\nabla_{W^t} J(X; \theta) \gets \left(C^t \righthook X^t\right)^* \cdot \left(\bignl_t'(Z^t) \odot (\bigpool_t^* \cdot e_y) \right)$ \Comment \eqref{eqn:dFdW:adjoint} with Thm. \ref{thm:Df:star}
\State $\nabla_{B^t} R(X; \theta) \gets \bignl_t'(Z^t) \odot \left(\bigpool^*_t \cdot e_w \right) + \bignl_t''(Z^t) \odot C^t(W^t, V^t) \odot \left(\bigpool^*_t \cdot e_v\right)$
\State $\nabla_{W^t} R(X; \theta) \gets \left(C^t \righthook X^t\right)^* \cdot \left(\bignl_t''(Z^t) \odot C^t(W^t, V^t) \odot \left(\bigpool_t^* \cdot e_v\right)\right)$
\State $\qquad + \left(C^t \righthook V^t\right)^* \cdot \left(\bignl_t'(Z^t) \odot \left(\bigpool_t^* \cdot e_v\right) \right) + \left(W^t \lefthook C^t\right)^* \cdot \left(\bignl'(Z^t) \odot \left(\bigpool_t^*\cdot e_w\right)\right)$
\State \Comment Both $\nabla_{B^t}R$ and $\nabla_{W^t}R$ can be computed via Thm. \ref{thm:main}, along with Thms. \ref{thm:Df:star} and \ref{thm:D_nab_f:star}
\State
\State $B^t \gets B^t - \eta \left(\nabla_{B^t} J(X; \theta) + \lambda \nabla_{B^t} R(X; \theta)\right)$
\State $W^t \gets W^t - \eta \left(\nabla_{W^t} J(X; \theta) + \lambda \nabla_{W^t} R(X; \theta)\right)$
\EndFor
\EndFunction
\end{algorithmic}
\end{algorithm}

\section{Conclusion and Future Work}
This work has developed a geometric framework for convolutional neural networks. The input data and parameters are defined over a vector space equipped with an inner product. The parameters are learned using a gradient descent algorithm that acts directly over the inner product space, avoiding the use of individual coordinates. Derivatives for higher-order loss functions are also explicitly calculated in a coordinate-free manner, providing the basis for a gradient descent algorithm. This mathematical framework can be extended to other types of deep networks, including recurrent neural networks, autoencoders and deep Boltzmann machines. Another interesting future direction is to expand the capabilities of automatic differentiation (AD) into this coordinate-free realm, strengthening the hierarchical approach to AD \cite{walter2013algorithmic}.

This paper has shown how to express a particular deep neural network, end-to-end, in a precise format. However, this framework should not be limited to only expressing previous results, and it should not be written off as simply a derivative calculation method. The stronger mathematical understanding of  neural networks provided by this work should promote expansion into new types of networks. 

\bibliographystyle{plain}

\begin{thebibliography}{1}

\bibitem{marsden1988manifolds}
R.~Abraham, J.~Marsden, and T.~Ratiu.
\newblock {\em Manifolds, Tensor Analysis, and Applications (2nd edition)}.
\newblock Springer, 1988.

\bibitem{werner1978multilinear}
W.~Greub.
\newblock {\em Multilinear Algebra}.
\newblock Springer Verlag, 1978.

\bibitem{huang2006large}
F.~Huang and Y.~LeCun.
\newblock Large-scale learning with {SVM} and convolutional networks for
  generic object recognition.
\newblock In {\em 2006 IEEE Computer Society Conference on Computer Vision and
  Pattern Recognition}, 2006.

\bibitem{krizhevsky2012imagenet}
A.~Krizhevsky, I.~Sutskever, and G.~Hinton.
\newblock Imagenet classification with deep convolutional neural networks.
\newblock In {\em Advances in neural information processing systems}, pages
  1097--1105, 2012.

\bibitem{lecun2015deep}
Y.~LeCun, Y.~Bengio, and G.~Hinton.
\newblock Deep learning.
\newblock {\em Nature}, 521(7553):436--444, 2015.

\bibitem{lecun1998gradient}
Y.~LeCun, L.~Bottou, Y.~Bengio, and P.~Haffner.
\newblock Gradient-based learning applied to document recognition.
\newblock {\em Proceedings of the IEEE}, 86(11):2278--2324, 1998.

\bibitem{rifai2011manifold}
S.~Rifai, Y.~Dauphin, P.~Vincent, Y.~Bengio, and X.~Muller.
\newblock The manifold tangent classifier.
\newblock In {\em Advances in Neural Information Processing Systems}, pages
  2294--2302, 2011.

\bibitem{simard1992tangent}
P.~Simard, B.~Victorri, Y.~LeCun, and J.~Denker.
\newblock Tangent {P}rop --- {A} formalism for specifying selected invariances
  in an adaptive network.
\newblock In {\em Advances in neural information processing systems}, pages
  895--903, 1992.

\bibitem{walter2013algorithmic}
S.~Walter and L.~Lehmann.
\newblock Algorithmic differentiation in {P}ython with {A}lgo{P}y.
\newblock {\em Journal of Computational Science}, 4(5):334--344, 2013.

\end{thebibliography}

\end{document}